\documentclass[letterpaper]{article} 
\usepackage{aaai25}  
\usepackage{times}  
\usepackage{helvet}  
\usepackage{courier}  
\usepackage[hyphens]{url}  
\usepackage{graphicx} 
\usepackage{natbib}  
\usepackage{caption} 
\usepackage{algorithm}
\usepackage{algorithmic}
\usepackage{graphicx}
\usepackage{subfigure}
\usepackage{amsmath}  
\usepackage{amsfonts} 
\usepackage{amssymb}  
\usepackage{newfloat}
\usepackage{listings}
\usepackage{xcolor}
\DeclareCaptionStyle{ruled}{labelfont=normalfont,labelsep=colon,strut=off} 
\floatstyle{ruled}
\newfloat{listing}{tb}{lst}{}
\floatname{listing}{Listing}

\title{Towards Automated Machine Learning Research}
\author{
Shervin Ardeshir
}
\affiliations{

}

\usepackage{bibentry}

\begin{document}

\maketitle

\begin{abstract}
This paper explores a top-down approach to automating incremental advances in machine learning research through component-level innovation, facilitated by Large Language Models (LLMs). Our framework systematically generates novel components, validates their feasibility, and evaluates their performance against existing baselines. A key distinction of this approach lies in how these novel components are generated. Unlike traditional AutoML and NAS methods, which often rely on a bottom-up combinatorial search over predefined, hardcoded base components, our method leverages the cross-domain knowledge embedded in LLMs to propose new components that may not be confined to any hard-coded predefined set. By incorporating a reward model to prioritize promising hypotheses, we aim to improve the efficiency of the hypothesis generation and evaluation process. We hope this approach offers a new avenue for exploration and contributes to the ongoing dialogue in the field.
\end{abstract}

%

\section{Introduction}

Efficient hypothesis generation, validation, and evaluation are critical, yet resource-intensive, components of scientific discovery. In many scientific fields, these processes require substantial manual effort, as they often involve intricate experiments and extensive data collection. The ability to streamline these tasks could significantly accelerate the pace of innovation.

Machine learning offers a unique opportunity in this regard. Unlike other scientific domains, hypothesis validation in machine learning can be automated through code, with effectiveness measured numerically using objective criteria such as loss or accuracy. This capability makes machine learning an ideal field for exploring automation in research.

Building on this potential, we propose a framework that leverages a top-down methodology using Large Language Models (LLMs) to generate high-level hypotheses. Although our approach is not intended to replace bottom-up methods such as AutoML-Zero\cite{real2020automl} or MetaQNN\cite{santoro2016meta}, it offers a complementary path by introducing cross-domain innovation and a broader exploration of potential solutions. By formulating and testing hypotheses in natural language, our method lowers the barrier to entry for a wider range of researchers, fostering interdisciplinary collaboration and the integration of diverse knowledge from various fields. This combination of top-down and bottom-up strategies improves the research pipeline, providing a more comprehensive and innovative approach to automated machine learning research.

In this paper, we contribute by:

\begin{itemize}
    \item \textbf{Proposing and evaluating viable components}: Generating viable hypotheses to replace neural network components and achieve competitive performance with known alternatives.
    \item \textbf{Training a reward model}: Learning patterns between the content of a hypothesis and its downstream performance.
    \item \textbf{Efficient Hypothesis Generation}: Using the reward model to prune and prioritize hypotheses, improving the efficiency of generation, validation, and evaluation.
\end{itemize}

\subsection{Caveats}

\begin{enumerate}
    \item This work does not make any assumptions about the inherent capabilities of LLMs to reason or have a deep understanding of ML topics. Even a random string generator can yield a meaningful hypothesis given unlimited attempts, akin to the infinite monkey theorem, which suggests that a monkey hitting keys at random on a typewriter for an infinite amount of time will almost surely type a given text, such as the complete works of Shakespeare. Our assumptions on the state of LLMs and ML are as follows.
    \begin{enumerate}
        \item LLMs are good enough at generating \textit{feasible} outputs, thus narrowing down our search space meaningfully from a set of random outputs.
        \item LLMs (and ML models in general) are good enough in pattern recognition. Therefore, training a reward model on performance would allow the model to identify common patterns among successful hypotheses.
    \end{enumerate}

    In short, we solely explore if LLMs can identify what would "look like" a good hypothesis based on the patterns that it has seen in previous examples.
    
    \item In this work, we solely lay the foundation and do not claim that our automated research necessarily would yield state-of-the-art (SOTA) results in machine learning research. We explore the feasibility of operationalizing steps involved in ML research to a level where the current state of LLMs can generate feasible hypotheses efficiently.
    
    \item This work was conducted using the authors' personal time and resources, limiting the scope to a small set of datasets and experiments. We hope that larger-scale experiments conducted at research labs interested in exploring this topic could further solidify this framework.
\end{enumerate}


\begin{figure}
    \centering
    \includegraphics[width=\linewidth]{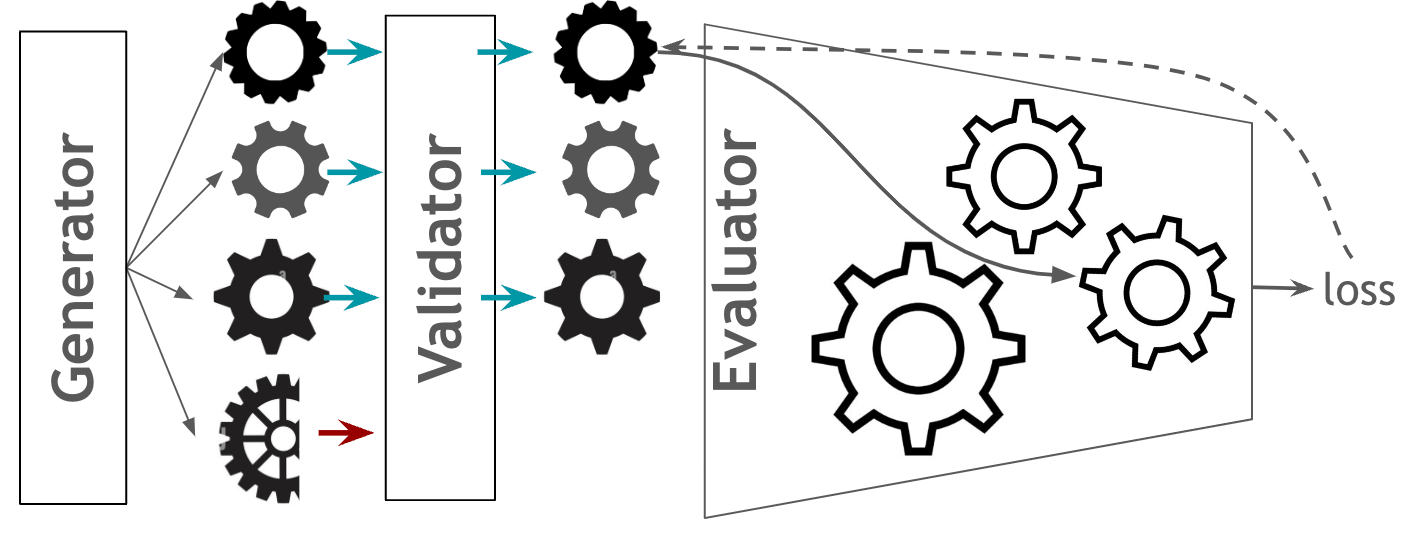}
    \caption{Framework overview: A set of hypotheses are generated to modify a specific component of an existing (hard-coded) framework. These hypotheses are tested for validity, evaluated for performance, and ranked, with the most promising candidates undergoing a full, computationally intensive evaluation.}
    \label{fig:teaser}
\end{figure}

This work assumes an existing baseline solution and explores incremental innovations in its components, focusing on one component at a time. For example, in a neural network, the component of interest might be an activation function. We generate a set of viable alternatives (hypotheses) and evaluate their performance against baseline components.

The framework includes a generator for hypothesis creation, a validator to ensure basic validity, and an evaluator to measure success metrics such as validation loss. A reward model is trained to prioritize hypotheses that perform well compared to baselines, reducing computational burden while maintaining a high probability of discovering valuable new hypotheses.

While we manually verified and adjusted the validator and evaluator functions, the generation of hypotheses was not manually reviewed, highlighting the potential for fully automated research. This framework aims to accelerate innovation and make advanced machine learning techniques more accessible, with potential applications in various scientific discovery tasks.

\section{Related Work}
The field of automated machine learning (AutoML)\cite{liang2019evolutionary} has rapidly advanced, automating processes such as data pre-processing, model selection, and hyperparameter tuning. Google's AutoML and Auto-Keras \cite{jin2019auto} have made machine learning more accessible. AutoML-Zero \cite{real2020automl} and MetaQNN \cite{baker2016designing} take a bottom-up combinatorial approach to model construction, evolving algorithms, and network architectures from basic operations. In contrast, our work uses a top-down method, leveraging large language models (LLMs) to start with high-level concepts, allowing for broader exploration and the potential for cross-domain innovation. 

\textbf{Meta-Learning} has also made significant strides, with foundational methods like MAML enabling fast task adaptation \cite{finn2017model}. Matching Networks \cite{vinyals2016matching} and Prototypical Networks \cite{snell2017prototypical} advanced few-shot learning, while optimization-based methods \cite{ravi2017optimization} and Memory-Augmented Neural Networks \cite{santoro2016meta} enhanced meta-learning capabilities. Simplified approaches like first-order meta-learning \cite{nichol2018first} and Probabilistic MAML \cite{finn2018probabilistic} further refined the field.

\textbf{Neural Architecture Search (NAS)} has progressed with approaches like NASNet \cite{zoph2018learning}, ENAS \cite{pham2018efficient}, and DARTS \cite{liu2018darts}, which introduced scalable and efficient architecture search methods. Auto-Keras \cite{jin2019auto} and ProxylessNAS \cite{cai2019proxylessnas} made NAS more accessible and practical, while AmoebaNet \cite{real2019regularized}, MnasNet \cite{tan2019mnasnet}, and FBNet \cite{wu2019fbnet} pushed the boundaries of mobile and hardware-aware optimization.

\textbf{Hyperparameter Optimization} has evolved with Bayesian optimization \cite{bergstra2011algorithms}, later improved by Snoek et al. \cite{snoek2012practical}. Random search \cite{bergstra2012random} provided a simpler alternative, while Hyperband \cite{li2017hyperband} and BOHB \cite{falkner2018bohb} optimized resource allocation. Gradient-based methods \cite{maclaurin2015gradient} and automated tuning for neural networks \cite{mendoza2016towards} further advanced the field.

The objective of our work aligns with ongoing efforts in AutoML, meta-learning, NAS, and hyperparameter optimization, however, it goes beyond those capabilities as our framework proposes and evaluates new components in a top-down approach and builds on the baseline state of the art.

\section{Framework}

The scope of this work begins when a specific area of machine learning research is selected, such as the development of a new activation function. The researcher selects this area and constructs a baseline set of solutions $\mathcal{B} = \{b_1, b_2, \ldots\}$ that represent the current state-of-the-art or commonly used approaches. The goal is to generate a set of viable alternatives $\mathcal{H} = \{h_1, h_2, \ldots\}$ to these baselines. Each proposed hypothesis $h_i$ is generated such that it can replace a component $b_j$ in the baseline solutions, such as substituting a new activation function in place of the standard ReLU in a neural network. This structured approach ensures that the generated hypotheses are directly relevant and potentially beneficial to the chosen area of research.

Our approach for generating and measuring the performance of each of the proposed hypotheses involves a generator, a validator, and an evaluator. A reward model is then trained to map the hypotheses to their success metrics measured by the evaluator. This reward is then used to improve the efficiency of the system by prioritizing more promising hypotheses solely by their content. In what follows we provide more details on each of these components. 

\subsection{The Generator}
\label{sec:generator}
The generator is the mechanism through which a feasible hypothesis $h_i$ is reached and sent for validation and evaluation. Here, it is a language model prompted by natural language, optionally followed by a reward model. In the activation function case study, these hypotheses take the form of novel activation functions. LLMs are trained on a comprehensive corpus of existing activation functions and related mathematical formulations, enabling them to propose viable functions.

We experiment with two types of base prompts. The first type encourages the model to discover incremental proposed blocks, to which we refer to as \textit{Incrementality Encouraging Prompting} (IEP for short). The following is an example of such a prompt, used for generating activation function blocks.

\textit{"define a python class that inherits from pytorch nn.Module. I should be able to use it as an activation function. Make sure if it has any parameters, all of them are set to default values so I can initialize without specifying any parameters. Try to come up with something that combines characteristics of Sigmoid/Tanh, ReLU, and ELU."}

The second type of base prompt aims to reduce the likelihood of trivial incrementality. We refer to this as \textit{Novelty Encouraged Prompting} (NEP) prompting. An example of such a prompt for activation function is as follows.

\textit{"define a python class that inherits from pytorch nn.Module. I should be able to use it as an activation function. Make sure if it has any params, all of them are set to default values so I can initialize without specifying any params. This function should not resemble common activation functions like ReLU, ELU, Sigmoid, or Tanh, and should explore unusual mathematical operations, combinations, or transformations. The expression can involve basic arithmetic, trigonometric functions, exponentials, or other non-linear operations, but avoid straightforward or commonly used forms in neural networks."}

The generator then involves a few wrappers around this base prompt, to request the implementation code for the proposed hypothesis in a parsable way.

Code \ref{code:activation_IEP} is an example of an activate function generated using the incrementality-encouraging base prompt. As prompted, the model clearly borrows characteristics from two commonly used activation functions of ReLU and Sigmoid.

\begin{lstlisting}[language=Python, 
                   caption=An example of an auto-generated activation function using Incrementality Encouraged Prompting (IEP), 
                   label=code:activation_IEP, 
                   frame=lines, 
                   numbers=none, 
                   basicstyle=\ttfamily\footnotesize, 
                   keywordstyle=\bfseries, 
                   stringstyle=\color{red}, 
                   showspaces=false, 
                   basicstyle=\ttfamily\scriptsize,
                   showstringspaces=false]
import torch
import torch.nn as nn
import torch.nn.functional as F
class HypothesisBlock(nn.Module):
    def __init__(self, alpha=1.0):
        super(HypothesisBlock, self).__init__()
        self.alpha = alpha    
    def forward(self, x):
        return torch.where(x >= 0, torch.sigmoid(x), self.alpha*(torch.exp(x) - 1))

\end{lstlisting}

Code \ref{code:activation_NEP} shows another activation function generated using the novelty-encouraged prompt. The model avoids directly using existing activation functions, adhering meaningfully to the instructions. While further exploration in prompt engineering is beyond this work’s scope, we believe it could significantly reduce LLMs' inductive bias towards state-of-the-art methods, minimize borrowing from existing literature, and encourage the proposal of less explored functions.

\begin{lstlisting}[language=Python, 
                   caption=An example of an auto-generated activation function using Novelty Encouraged Prompting (NEP), 
                   label=code:activation_NEP, 
                   frame=lines, 
                   numbers=none, 
                   basicstyle=\ttfamily\footnotesize, 
                   keywordstyle=\bfseries, 
                   stringstyle=\color{red}, 
                   basicstyle=\ttfamily\scriptsize,
                   showspaces=false, 
                   showstringspaces=false]
import torch
import torch.nn as nn
class HypothesisBlock(nn.Module):
    def __init__(self, scale=1.0, offset=0.1):
        super(HypothesisBlock, self).__init__()
        self.scale = scale
        self.offset = offset
    def forward(self, x):
        return self.scale * torch.sin(x) * torch.exp(-torch.abs(x)) + self.offset
\end{lstlisting}

\begin{figure}[ht]
    \centering
    \begin{minipage}{0.23\textwidth}
        \centering
        \includegraphics[width=\textwidth]{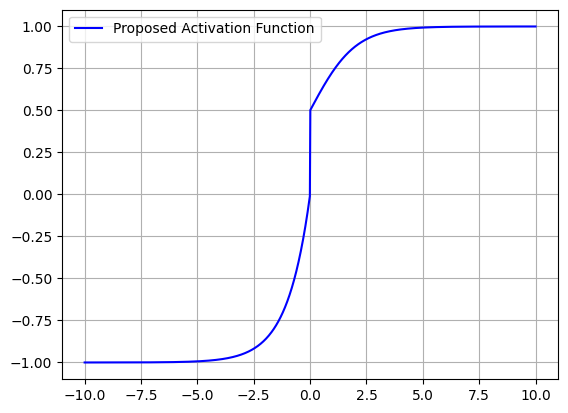}
        \label{fig:incremental_activation}
    \end{minipage}
    \begin{minipage}{0.23\textwidth}
        \centering
        \includegraphics[width=\textwidth]{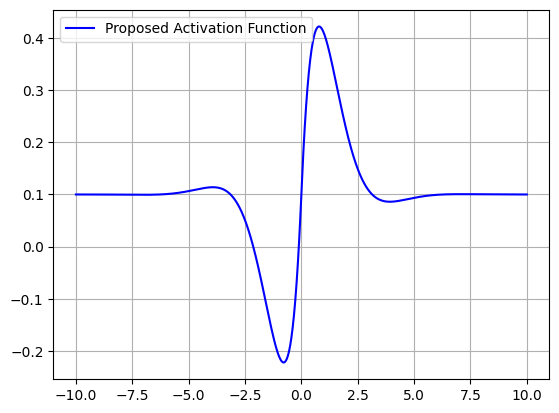}
        \label{fig:novel_activation}
    \end{minipage}
    \caption{Activation functions generated through Incrementality Encouraging (left) and Novelty Encouraging (right) prompting.}
    \label{fig:incremental_novel_activation}
\end{figure}

Figure 
\ref{fig:incremental_novel_activation} visualizes the shape of the proposed activation functions.

For more examples, please refer to the Appendix.

\subsection{The Validator}

Each proposed hypothesis is first passed through a validator function, denoted $v$. This function checks the validity of the hypothesis, ensuring it meets necessary criteria before further evaluation. For a proposed activation function $h_i$, the validator function $v(h_i)$ returns a binary value indicating whether $h_i$ is a valid activation function (that is, $v(h_i) \in \{0, 1\}$).

For our case study on activation functions, we implement the validator manually and as a unittest\footnote{Fully automating this function is very feasible, but out of the scope of this effort.}. The validator checks if $h_i$ inherits from nn.Module, has the required initialization and forward functions, all its parameters have default values, and can pass a few basic test cases. We provide code snippets of this validator in the Appendix section Code \ref{code:validator_activation}.

\subsection{The Evaluator}

Valid hypotheses are fully evaluated using an evaluation function, denoted $e$. The goal of this function is to replace a baseline component $b_j$ with an alternative hypothesis $h_i$ and measure the performance of the model in the task(s) of interest. For our case, this translates to integrating the hypothesis into a machine learning model and measuring its performance, specifically its loss on the validation split of the dataset (val-loss). To streamline the process, we perform a single iteration of forward and backward passes to obtain a preliminary loss value, rather than conducting a full training and evaluation cycle. Formally, for a model $m$ and an activation function $h_i$, the val loss is calculated as $e(m(h_i))$.

In our experiments, we hard-coded the architecture to a 2-layer MLP, and used cross-entropy and MSE loss for a set of classification and regression tasks. We also define the problem as solving the classification and regression instances in a one-pass learning setup (only one epoch) to reduce the computation required for each $h_i$ and enable more extensive exploration across a larger set of hypotheses.

\subsection{The Reward}
\label{sec:reward}
Passing the set of hypotheses $\mathcal{H}$ to the evaluator function results in the collection of pairs of generated hypotheses (activation functions) and their corresponding validation losses, in the form of $(h_i, l(h_i))$. We define the reward as the win rate of the proposed hypothesis $h_i$ over the baselines. Specifically, we measure two metrics:

\textbf{Baseline Win Rate (B-WR):}
This metric measures the percentage of times $h_i$ outperforms any given baseline $b_j$ across different tasks and over different runs. Formally, it measures $P(l(h_i) < l(b_j))$.

\textbf{Baseline State-of-the-Art Win Rate (BSOTA-WR):}
This metric measures the win rate of the proposed hypothesis over the best runs of the entire baseline set $\mathcal{B}$ in each task / dataset. Formally, it is defined as $P(l(h_i) < \min(l(b_j))|_{j=1,\ldots,|\mathcal{B}|})$. Please note that the minimum operation is done on the average loss of different runs for each baseline, thus the best baseline run still contains a distribution of losses (resulting from several runs / random initialization), allowing for calculating a probabilistic win rate.

The reward model is then trained as a ranking model mapping the content of the hypothesis $h_i$ to its downstream performance (i.e. loss). In other words, this model is aimed to looking at the content of a proposed component (i.e. code of an activation function), and be able to predict how well it is likely to perform in terms of winning over the baseline set. Intuitively, the reward model learns patterns in the content of the proposed activation functions, leading to better performance.

\subsection{Closing the Loop}

In the initial round of hypothesis generation, every hypothesis is evaluated using a brute-force approach, where each one is passed through the validator and evaluator in a fully exploratory iteration. This process allows for comprehensive data collection, yielding the success metrics B-WR and BSOTA-WR for each validated and evaluated hypothesis. We employ three LLMs to generate a dataset of 2000 validated and evaluated hypotheses for each component type.

Once this initial data is collected, the second iteration leverages a trained reward model to streamline the process. The reward model is used to prune the newly generated hypotheses, filtering them to select the top-$k$ candidates based on their predicted performance. These top-$k$ hypotheses, expected to be the most promising, are the only ones that proceed to the full evaluation phase, which involves more intensive computational resources.






\section{Experiments}

We aim to identify novel components that improve a simple neural network's performance across various tasks using a one-pass learning setup, where the model is trained for a single epoch. This approach enables rapid iteration and evaluation of numerous hypotheses.

\subsection{Experimental Setup}

\subsubsection{Downstream Tasks and Datasets}
To validate the effectiveness of our framework, we performed experiments on six tasks using four well-known datasets, covering both classification and regression. 

\textbf{Iris Dataset:} Classification task with 150 instances, 4 features, and 3 classes.

\textbf{Wine Dataset:} Used for both classification (3 classes) and regression, with 178 instances and 13 features.

\textbf{Breast Cancer Dataset:} Binary classification task with 569 instances and 30 features.

\textbf{Diabetes Dataset:} Regression task with 442 instances, each with 10 baseline variables, such as age, sex, body mass index (BMI), average blood pressure, and six blood serum measurements. The goal is to predict the progression of the disease one year after baseline.

\subsubsection{Generated Hypothesis Dataset}
Each hypothesis is evaluated to generate a dataset of 36,000 (hypothesis, reward) tuples over two iterations of 18,000 each. These are further divided into 3 LLMs, 3 component types, and 2 prompt types, with each combination generating 1,000 samples.

\subsubsection{Components}
We experimented with three component types: activation functions, regularization functions, and preprocessing functions. Detailed prompts for each type are provided in the appendix.

\subsubsection{Language Models and Prompts}

We used three language models: GPT-3.5 Turbo, GPT-4o and Gemini, to generate components, using two prompt types: incrementality-encouraging and novelty-encouraging (as described in Section \ref{sec:generator}).

\subsubsection{Architecture}
We employed a 2-layer fully-connected neural network with 64 and 16 units for all datasets for simplification.

\subsection{Quantitative Results and Analysis}
\label{sec:results}

Our goal is to generate \textbf{viable} and \textbf{high-performing} proposed components, \textbf{efficiently}. In the following, we provide details on how we measure success in these aspects.  

\subsubsection{Performance: Component Evaluation}
As mentioned in the Reward Section \ref{sec:reward}, we use the two key metrics of Baseline Win-rate (B-WR for short), and Baseline State-of-the-art win-rate (BSOTA-WR for short) to assess the effectiveness of each proposed block. Both metrics, BSOTA-WR and B-WR, are designed to provide a holistic view of the proposed method's performance, highlighting its potential to advance the state of the baseline set by setting new benchmarks and consistently outperforming the baseline set. Table. \ref{tab:component_metrics}, contains the metrics for the set of components generated through the pipeline. We also report the success rate of the Validator, indicating the fraction of generated hypotheses that had the valid format. Please note that all of these metrics are averaged over the whole dataset of hypotheses generated in the first iteration (2000 samples generated for each component type). The performance of the individual components can be seen in the scatter plots provided in Figure. \ref{fig:win_rate_scatter_activation}. As can be observed, in the majority of cases the generated hypotheses have a low win rate compared to the baseline set, however, there are components with win rates very close to 1, indicating that they always outperform every single baseline individually and also the baseline state of the art.

\begin{figure}
    \centering
    \includegraphics[width=.9\linewidth]{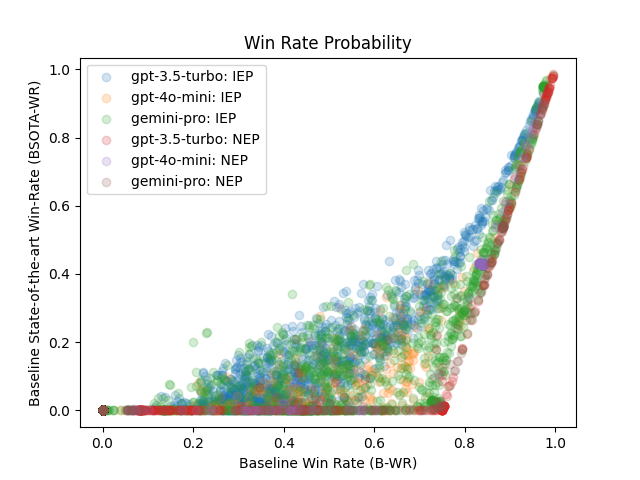}
    \caption{Scatter Plots of Win Rate Probabilities: These scatter plots illustrate the relationship between the Baseline Win Rate (B-WR) and the Baseline State-of-the-Art Win Rate (BSOTA-WR) for several hypotheses across different blocks. The y-axis represents the BSOTA-WR, while the x-axis represents the B-WR. By definition, a hypothesis reaching 1 in one axis will reach 1 in the other. Also, as expected, it can be seen from the distributions that BSOTA-WR is generally a more difficult objective to achieve.}
    \label{fig:win_rate_scatter_activation}
\end{figure}

\begin{table}[h]
\centering
\resizebox{\columnwidth}{!}{%
\begin{tabular}{|c|c|c|c|}
\hline
- & Inc & Nov \\ \hline
Activation & \begin{tabular}[c]{@{}c@{}}Validator-PR: 0.9756\\ B-WR: 0.0474\\ BSOTA-WR: 0.0107\end{tabular} & \begin{tabular}[c]{@{}c@{}}Validator-PR: 0.4340\\ B-WR: 0.5150\\ BSOTA-WR: 0.2501\end{tabular} \\ \hline
Preprocessor & \begin{tabular}[c]{@{}c@{}}Validator-PR: 0.9175\\ B-WR: 0.0929\\ BSOTA-WR: 0.3920\end{tabular} & \begin{tabular}[c]{@{}c@{}}Validator-PR: 0.9664\\ B-WR: 0.2167\\ BSOTA-WR: 0.3883\end{tabular} \\ \hline
Regularizer & \begin{tabular}[c]{@{}c@{}}Validator-PR: 0.83127\\ B-WR: 0.3726\\ BSOTA-WR: 0.3047\end{tabular} & \begin{tabular}[c]{@{}c@{}}Validator-PR: 0.8678\\ B-WR: 0.6272\\ BSOTA-WR: 0.5860\end{tabular} \\ \hline
\end{tabular}%
}
\caption{Comparing validator passing rate and evaluator metrics based on the two types of prompting, for each component type.}
\label{tab:component_metrics}
\end{table}

\subsubsection{Reward model Evaluation}
As mentioned earlier, the goal of the reward model is to learn to predict the performance of a proposed hypothesis solely from its content (code). To train such a model, we extract three different code embedding features from the content of the implementation code generated, namely CodeBERT\cite{codebert}, GraphCodeBert\cite{graphcodebert}, and CodeGen\cite{codegen}. We report the results on ranking models trained on the concatenation of all three features in table \ref{tab:reward_metrics_activation}. In the Appendix, we also provide ablation on the same metrics for each feature type, and also for preprocessors and regularizers. 

We use established ranking metrics, including Kendall's Tau ($k-\tau$), Spearman correlation coefficient (SCC), and Pearson correlation coefficient (PCC), as reported in Tables \ref{tab:reward_metrics_activation}, \ref{tab:reward_metrics_preprocessing}, and \ref{tab:reward_metrics_activation} for activation functions, preprocessing functions, and regularization functions respectively. From table \ref{tab:reward_metrics_activation}, for activation, preprocessing, and regularization functions, respectively. These metrics show successful generalization of the ranking models across components generated by different language models. While reward models perform best on the datasets they were trained on, the consistently positive correlations across different language models demonstrate their robustness and generalization capability. Even when correlation values are modest, they remain directionally positive, indicating meaningful ranking performance.

\begin{table*}[ht]
\centering
\begin{tabular}{|c|c|c|c|}
\hline
 train test & \textbf{gpt-3.5-turbo} & \textbf{gpt-4o-mini} & \textbf{gemini-pro} \\
\hline
\textbf{gpt-3.5-turbo} & \begin{tabular}{c}
k-$\tau$: (0.824, 0.000) \\
SCC: (0.922, 0.000) \\
PCC: (0.931, 0.000)
\end{tabular} & \begin{tabular}{c}
k-$\tau$: (0.627, 0.000) \\
SCC: (0.805, 0.000) \\
PCC: (0.754, 0.000)
\end{tabular} & \begin{tabular}{c}
k-$\tau$: (0.653, 0.000) \\
SCC: (0.814, 0.000) \\
PCC: (0.780, 0.000)
\end{tabular} \\
\hline
\textbf{gpt-4o-mini} & \begin{tabular}{c}
k-$\tau$: (0.284, 0.000) \\
SCC: (0.352, 0.000) \\
PCC: (0.426, 0.000)
\end{tabular} & \begin{tabular}{c}
k-$\tau$: (0.471, 0.000) \\
SCC: (0.527, 0.000) \\
PCC: (0.792, 0.000)
\end{tabular} & \begin{tabular}{c}
k-$\tau$: (0.122, 0.000) \\
SCC: (0.153, 0.000) \\
PCC: (0.196, 0.000)
\end{tabular} \\
\hline
\textbf{gemini-pro} & \begin{tabular}{c}
k-$\tau$: (0.315, 0.000) \\
SCC: (0.428, 0.000) \\
PCC: (0.410, 0.000)
\end{tabular} & \begin{tabular}{c}
k-$\tau$: (0.248, 0.000) \\
SCC: (0.337, 0.000) \\
PCC: (0.310, 0.000)
\end{tabular} & \begin{tabular}{c}
k-$\tau$: (0.505, 0.000) \\
SCC: (0.672, 0.000) \\
PCC: (0.633, 0.000)
\end{tabular} \\
\hline
\end{tabular}
\caption{Reward model performance across different datasets on the activation function block. Please find similar tables for the other blocks (pre-processor and regularization function) in the appendix.}
\label{tab:reward_metrics_activation}
\end{table*}

\subsubsection{Efficienct Hypothesis Evaluation}
We also evaluate the efficiency of the reward model in terms of prioritizing the candidates in the second iteration. That means that we measure the performance of the top 50 candidates at each step if we were to sort the candidates based on their predicted reward. Intuitively, a good reward model would sort them in an order in which the top k candidates are more likely to be on the top of the list, therefore discovering the promising hypotheses earlier, resulting in a curve with a higher AUC. Figure. \ref{fig:activation_efficiency} visualize the efficiency curves for the activation function on the datasets generated by the different LLMs separately. We provide similar curves for the datasets generated for other components (preprocessor and regularizer) in Figures \ref{fig:preprocessor_efficiency} and \ref{fig:regularizer_efficiency} of the appendix. The x-axis in these figures shows the number of steps, and the y-axis is the reward (linear addition of BSOTA-WR + B-WR) for the top 50 hypotheses if they were to be prioritized by the reward model of interest. In all graphs, the blue curve shows how fast the pipeline reaches the high top-50 accuracies if there is no reward model used (chance/random reward). As it can be observed, all reward models for the activation functions lead to a faster discovery of better hypotheses, leading to higher AUC values. 
\begin{figure*}[ht]
    \centering
    \begin{minipage}{0.33\textwidth}
        \centering
        \includegraphics[width=\textwidth]{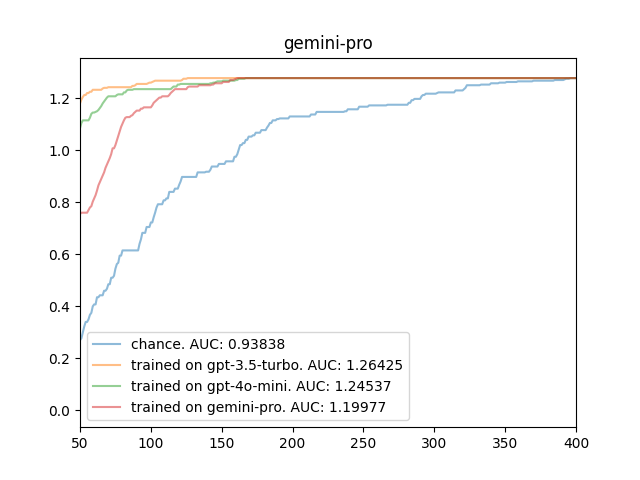}
        \label{fig:image1}
    \end{minipage}
    \begin{minipage}{0.33\textwidth}
        \centering
        \includegraphics[width=\textwidth]{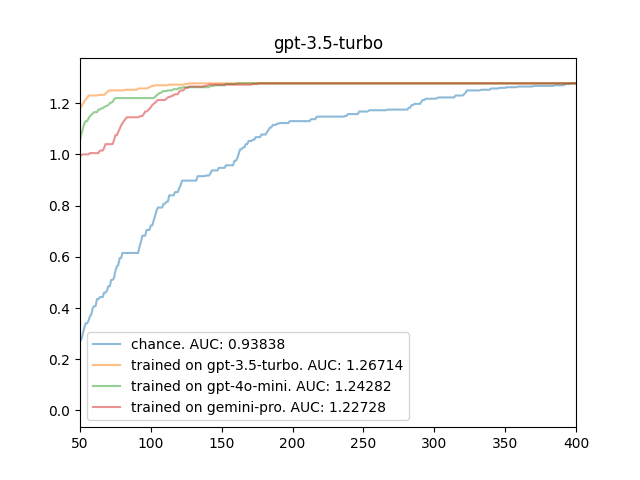}
        \label{fig:image2}
    \end{minipage}
    \begin{minipage}{0.33\textwidth}
        \centering
        \includegraphics[width=\textwidth]{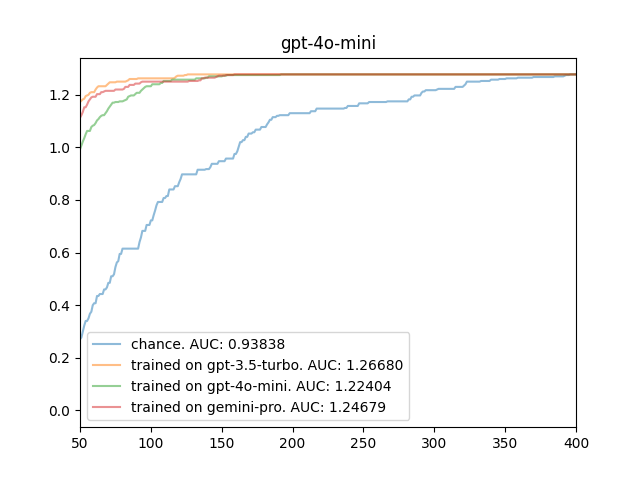}
        \label{fig:image3}
    \end{minipage}
    \caption{Efficiency for the activation function Reward Models. Each graph illustrates how efficiently the respective reward model prioritizes high-performing hypotheses, under different scenarios of being trained and tested on different LLM-generated hypothesis dataset.}
    \label{fig:activation_efficiency}
\end{figure*}


\section{Risks and Limitations}
Given that this work is the first in the lane. here we cover some potential risks and limitations for this line of research. These risks may be even more prominent once models are trained or fine-tuned end-to-end in a closed-loop setup with minimal human involvement.

\textbf{Shortcuts:} Given the empirical nature of this approach, there's a possibility that the model might exploit existing shortcuts rather than genuinely innovative solutions. This could lead to overfitting to specific datasets or tasks without contributing to broader advancements.

\begin{figure}
    \centering
    \includegraphics[width=\linewidth]{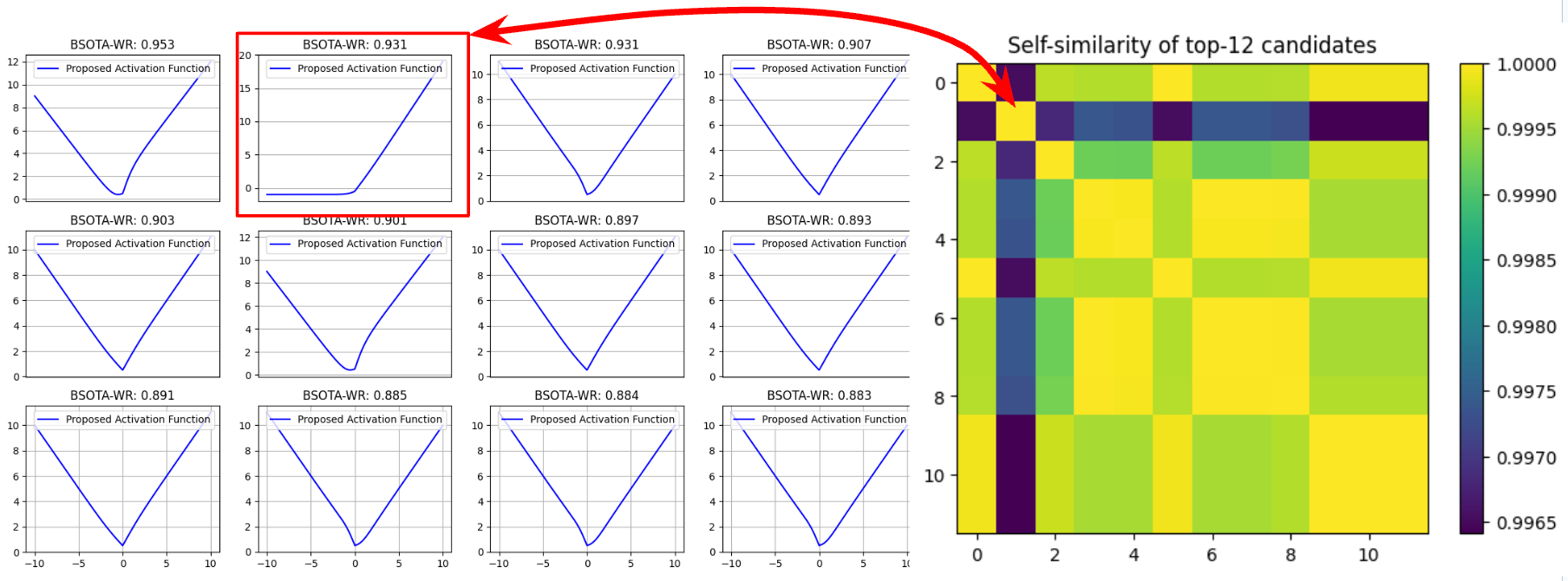}
    \caption{Left Panel: Graphical representation of the top-12 proposed activation functions in one of the runs. Each graph shows the shape of the activation function along with its Baseline State-Of-The-Art Win Rate (BSOTA-WR), and novelty score (N-score). The highlighted function (in red) demonstrates a notable balance between performance (BSOTA-WR: 0.931) and novelty (N-score: 0.034). Right Panel: Self-similarity heatmap of the top-12 activation functions. The color scale represents the degree of similarity, with yellow indicating high similarity and blue indicating lower similarity. This matrix helps to identify clusters of similar functions, highlighting the uniqueness of each proposed activation function. }
    \label{fig:top_12_before}
\end{figure}

\textbf{Reward Collapse:} During our experiments, we observed a significant issue with the generation of redundant and highly similar hypotheses, particularly when using GPT-3.5-turbo to generate activation functions. As illustrated in Figure \ref{fig:top_12_before}, the top-12 activation functions often exhibit striking similarities, indicating a lack of diversity in the generated hypotheses. This phenomenon, known as reward collapse, occurs when the reward model becomes overly focused on specific patterns, leading to a narrow exploration of the hypothesis space. The right side of the figure, shows the pairwise similarity between the top-12 candidates, it can be observed that the one difference component (highlighted in red) completely stands out both in terms of the shape of its activation function, and also in terms of its similarity to others in the embeddings space. Given this phenomenon, we did a preliminery exploration on whether we can construct a set of diverse activation functions by constructing a set iteratively and greedily as a trade-off of win rate and diversity. 

\begin{figure}
    \centering
        \includegraphics[width=\linewidth]{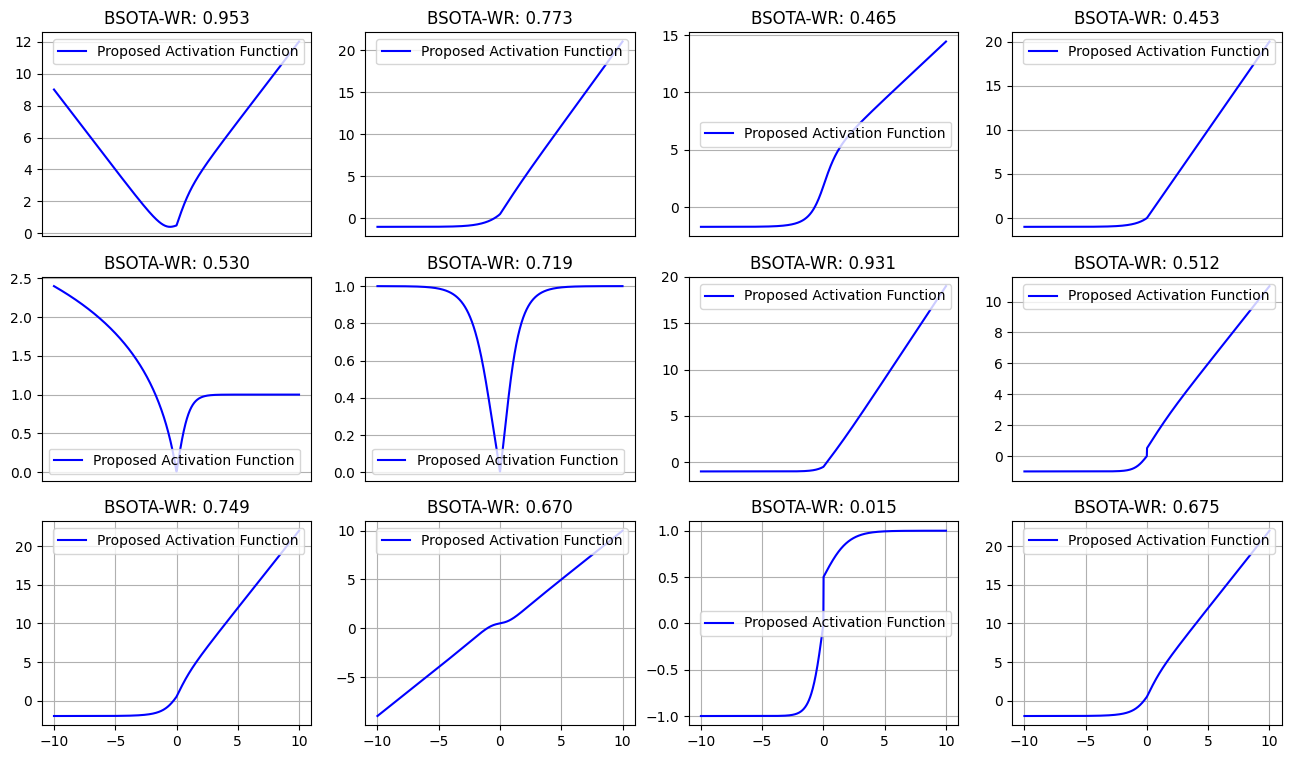}
    \caption{Top 12 activation functions generated after applying the greedy algorithm for balancing performance and diversity. The figure shows a diverse set of activation functions selected through an iterative process that maximizes both win rate and embedding distance from previously selected functions. This method mitigates redundancy and encourages the exploration of innovative and varied solutions, as reflected in the distinct characteristics of the top 12 functions.}

    \label{fig:top_12_after}
\end{figure}

This greedy and iterative approach encourages the selection of hypotheses that are both high-performing and diverse, thereby promoting a broader and more thorough exploration of the hypothesis space. By balancing the exploitation of known successful solutions with the exploration of novel and potentially superior alternatives, this method helps mitigate the risk of reward collapse. The effectiveness of this approach is demonstrated in Figure \ref{fig:top_12_after}, where the top-12 activation functions constructed with this method exhibit a greater diversity compared to the initial set. This suggests the possibility of preventing collapse in case of finetuning the generator (future work).

\textbf{SOTA Inductive Bias and unintended plagiarism:}
LLMs, trained on vast datasets, risk generating outputs that closely resemble existing works, leading to unintended plagiarism and a bias toward state-of-the-art (SOTA) methodologies. This limits innovation, as models may favor incremental changes over novel ideas. To address this, it's essential to build careful baseline sets and implement strong credit assignment. Prompt design also plays a key role; novelty-focused prompts yield more diverse outputs, while those targeting incremental improvements often mirror existing literature. Refining prompts to avoid reliance on known solutions and explore new areas can reduce plagiarism and SOTA bias, encouraging truly innovative contributions.

\textbf{Optimizing for incremental short term improvements:} The empirical focus of this work, combined with the absence of strong theoretical constraints, creates a risk of prioritizing short-term, incremental gains over more significant, long-term advancements. This approach may lead to the discovery of surface-level improvements that offer marginal benefits, while potentially overlooking opportunities for groundbreaking innovations that could drive substantial progress in the field.

\section{Future Work}

Future research could focus on expanding the framework to other types of machine learning components beyond activation functions, preprocessors and regularizers, and including more complex architectures and diverse datasets. Additionally, refining the reward model to balance novelty and performance more effectively, and incorporating stronger theoretical constraints, could help mitigate risks like reward collapse and incremental bias. An intriguing direction for future work is fine-tuning the language model based on the reward signal, which could guide the model towards generating higher-quality and more innovative hypotheses. Moreover, ensuring that the embeddings extracted from the generated hypotheses are consistent with those of the backbone model could open the possibility for fully differentiable training, further enhancing the integration and efficiency of the framework. Further experiments could also explore constrcuting prompt engineering practices to reduce unintended plagiarism and inductive bias. And last but not least, a proper credit assignment framework would be a necessity for improviong this line of research.

\section{Conclusion}

This work introduces a framework for automating machine learning research by leveraging large language models to generate, validate, and evaluate novel components. While the approach shows promise in enhancing the efficiency of hypothesis generation and evaluation, it also presents challenges, such as the risk of reward collapse and the tendency to prioritize incremental improvements. Addressing these issues through careful design, fine-tuning strategies, and future refinements—such as consistent embedding integration for fully differentiable training—will be key to realizing the full potential of this automated research paradigm.

\bibliography{main}

\clearpage
\section{Appendix}
\subsection{Performance of the generated components}
\begin{figure*}[ht]
    \centering
    \begin{minipage}{0.33\textwidth}
        \centering
        \includegraphics[width=\textwidth]{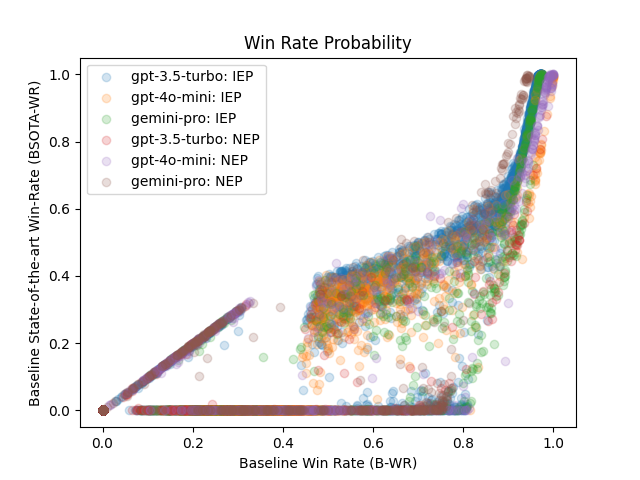}
        \caption{Preprocessor}
        \label{fig:win_rate_scatter_preprocessor}
    \end{minipage}
    \begin{minipage}{0.33\textwidth}
        \centering
        \includegraphics[width=\textwidth]{figures/BWR_BSOTA_scatter_activation.png}
        \caption{Activation Function}
        \label{fig:win_rate_scatter_activation}
    \end{minipage}
    \begin{minipage}{0.33\textwidth}
        \centering
        \includegraphics[width=\textwidth]{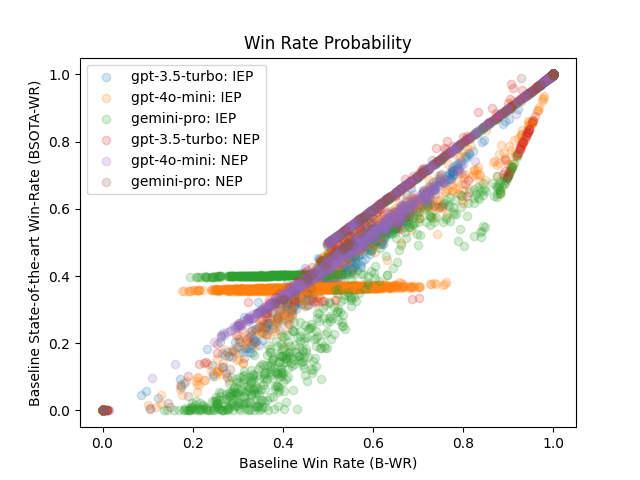}
        \caption{Regularizer}
        \label{fig:win_rate_scatter_regularizer}
    \end{minipage}
    \caption{Scatter Plots of Win Rate Probabilities: These scatter plots illustrate the relationship between the Baseline Win Rate (B-WR) and the Baseline State-of-the-Art Win Rate (BSOTA-WR) for several hypotheses across different blocks. The y-axis represents the BSOTA-WR, while the x-axis represents the B-WR. By definition, a hypothesis reaching 1 in one axis will reach 1 in the other. Also, it can be seen from the distributions that BSOTA-WR is generally a more difficult objective to achieve.}
    \label{fig:win_rate_scatter_combined}
\end{figure*}

As mentioned in section \ref{sec:results} we present a scatter plot visualization to compare the two success metrics for the hypotheses generated across different block types. Figure \ref{fig:win_rate_scatter_combined} illustrates the scatter plots for the pre-processor, activation, and regularization block types, respectively.

\subsection{Reward Model Efficiency}
\begin{figure*}[ht]
    \centering
    \begin{minipage}{0.33\textwidth}
        \centering
        \includegraphics[width=\textwidth]{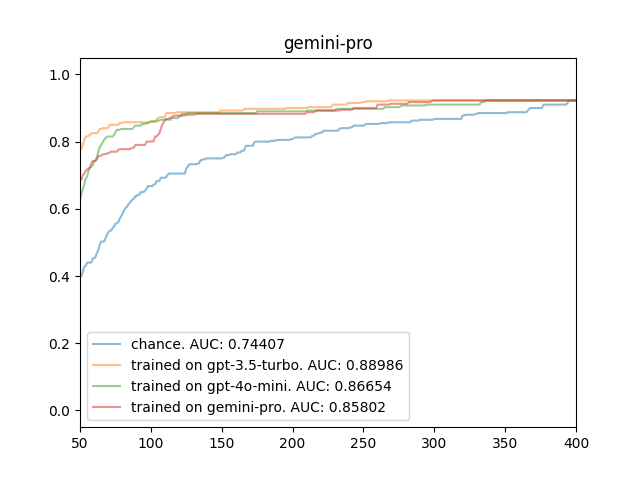}
        \label{fig:image1}
    \end{minipage}
    \begin{minipage}{0.33\textwidth}
        \centering
        \includegraphics[width=\textwidth]{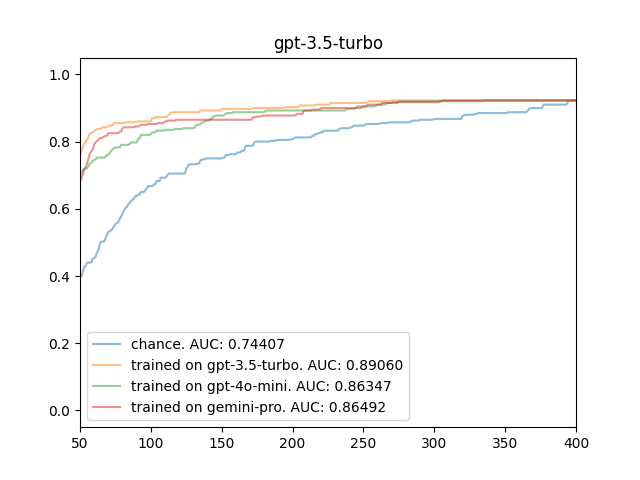}
        \label{fig:image2}
    \end{minipage}
    \begin{minipage}{0.33\textwidth}
        \centering
        \includegraphics[width=\textwidth]{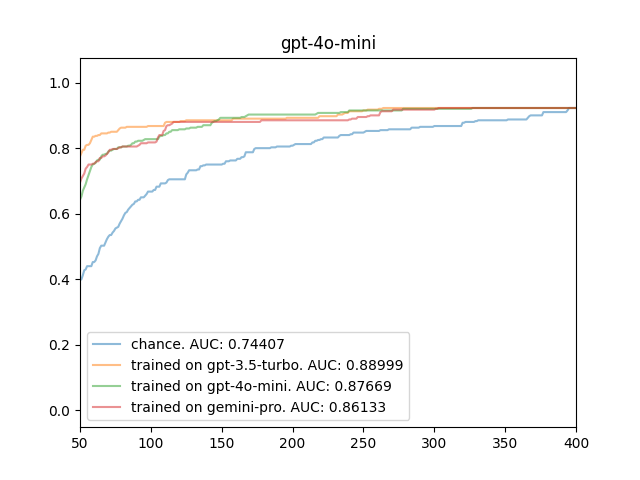}
        \label{fig:image3}
    \end{minipage}
    \caption{Efficiency for Preprocessor Reward Models.}
    \label{fig:preprocessor_efficiency}
\end{figure*}

\begin{figure*}[ht]
    \centering
    \begin{minipage}{0.33\textwidth}
        \centering
        \includegraphics[width=\textwidth]{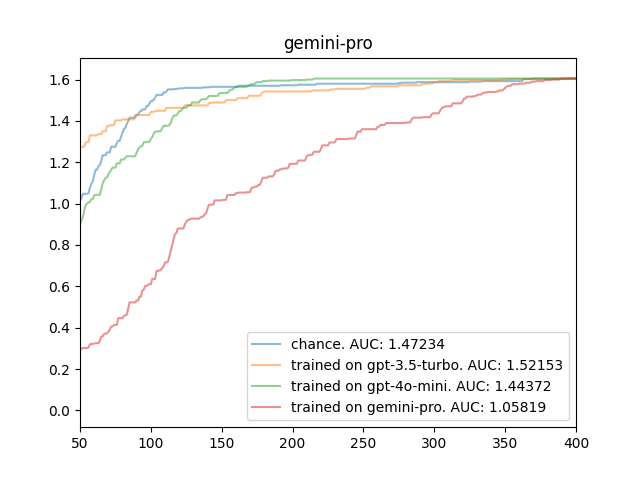}
        \label{fig:image1}
    \end{minipage}
    \begin{minipage}{0.33\textwidth}
        \centering
        \includegraphics[width=\textwidth]{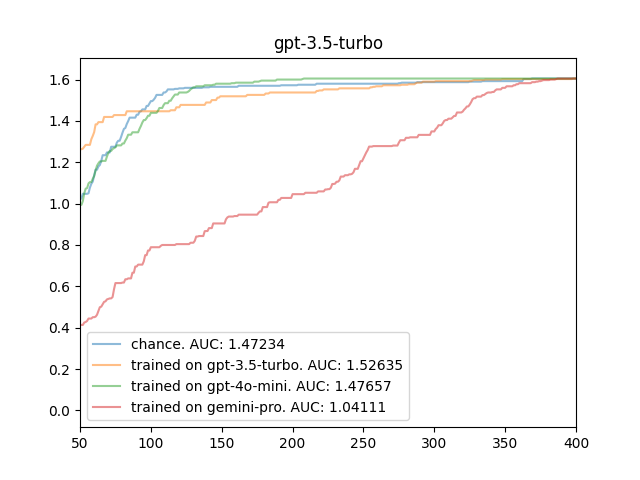}
        \label{fig:image2}
    \end{minipage}
    \begin{minipage}{0.33\textwidth}
        \centering
        \includegraphics[width=\textwidth]{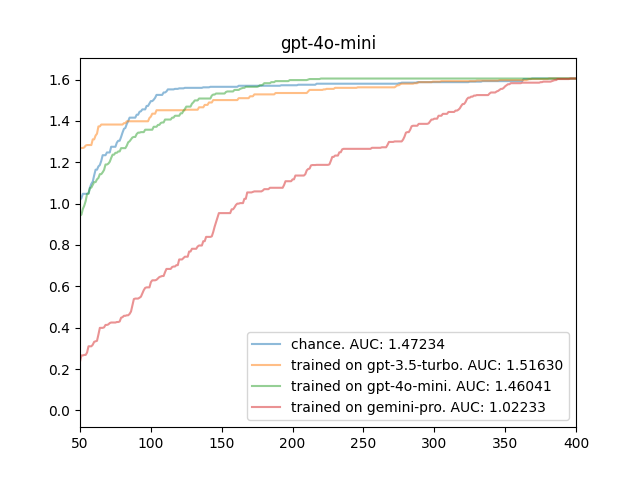}
        \label{fig:image3}
    \end{minipage}
    \caption{Efficiency for regularizer function Reward Models.}
    \label{fig:regularizer_efficiency}
\end{figure*}

Here we provide efficiency curves for the preprocessor and regularize blocks in figure \ref{fig:preprocessor_efficiency} and \ref{fig:regularizer_efficiency} respectively. It can be observed that similar to activation functions, the reward ranking model trained on the preprocessor blocks can effectively speed up the discovery of the most promising proposed components. However, when it comes to the regularizers, the trained reward models, especially the ones trained on the Gemini-pro dataset, fail to generalize to other datasets. We also provide the metrics for the rewards models in tables \ref{tab:reward_metrics_preprocessing} and \ref{tab:reward_metrics_regularizer} respectively. 

\begin{figure*}[ht]
    \centering
    \begin{minipage}{0.33\textwidth}
        \centering
        \includegraphics[width=\textwidth]{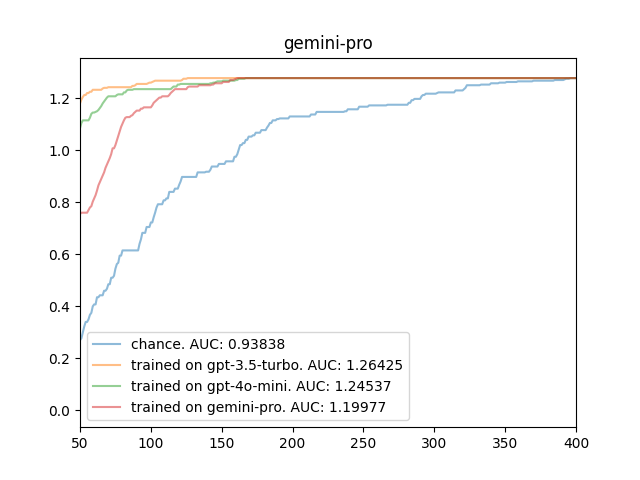}
        \label{fig:image1}
    \end{minipage}
    \begin{minipage}{0.33\textwidth}
        \centering
        \includegraphics[width=\textwidth]{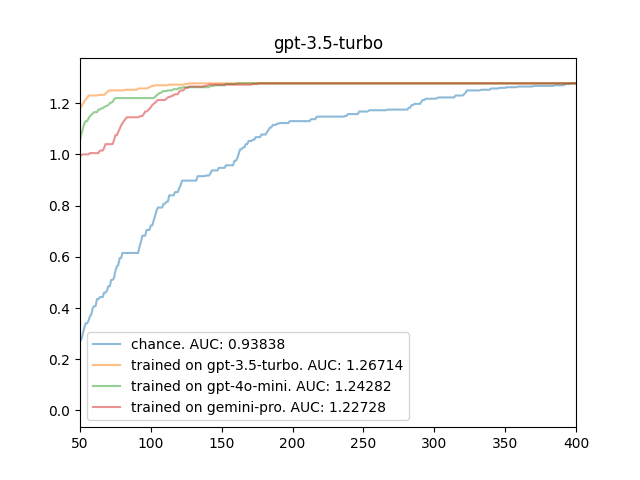}
        \label{fig:image2}
    \end{minipage}
    \begin{minipage}{0.33\textwidth}
        \centering
        \includegraphics[width=\textwidth]{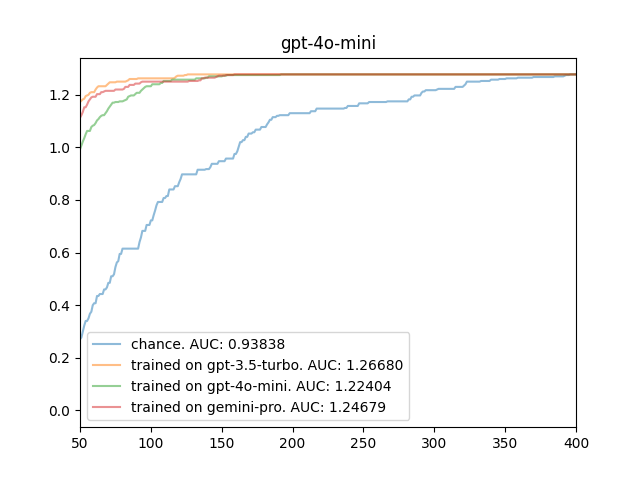}
        \label{fig:image3}
    \end{minipage}
    \caption{Efficiency for activation function Reward Models.}
    \label{fig:activation_efficiency}
\end{figure*}

\begin{table*}[ht]
\centering
\begin{tabular}{|c|c|c|c|}
\hline
 train \ test  & \textbf{gpt-3.5-turbo} & \textbf{gpt-4o-mini} & \textbf{gemini-pro} \\
\hline
\textbf{gpt-3.5-turbo} & \begin{tabular}{c}
k-$\tau$: (0.606, 0.000) \\
SCC: (0.740, 0.000) \\
PCC: (0.773, 0.000)
\end{tabular} & \begin{tabular}{c}
k-$\tau$: (0.405, 0.000) \\
SCC: (0.550, 0.000) \\
PCC: (0.521, 0.000)
\end{tabular} & \begin{tabular}{c}
k-$\tau$: (0.432, 0.000) \\
SCC: (0.600, 0.000) \\
PCC: (0.645, 0.000)
\end{tabular} \\
\hline
\textbf{gpt-4o-mini} & \begin{tabular}{c}
k-$\tau$: (0.252, 0.000) \\
SCC: (0.346, 0.000) \\
PCC: (0.271, 0.000)
\end{tabular} & \begin{tabular}{c}
k-$\tau$: (0.395, 0.000) \\
SCC: (0.524, 0.000) \\
PCC: (0.592, 0.000)
\end{tabular} & \begin{tabular}{c}
k-$\tau$: (0.327, 0.000) \\
SCC: (0.442, 0.000) \\
PCC: (0.459, 0.000)
\end{tabular} \\
\hline
\textbf{gemini-pro} & \begin{tabular}{c}
k-$\tau$: (0.039, 0.486) \\
SCC: (0.066, 0.403) \\
PCC: (0.105, 0.180)
\end{tabular} & \begin{tabular}{c}
k-$\tau$: (0.139, 0.013) \\
SCC: (0.193, 0.013) \\
PCC: (0.203, 0.009)
\end{tabular} & \begin{tabular}{c}
k-$\tau$: (0.298, 0.000) \\
SCC: (0.403, 0.000) \\
PCC: (0.415, 0.000)
\end{tabular} \\
\hline
\end{tabular}
\caption{Reward model performance across different datasets on the pre-processing function block}
\label{tab:reward_metrics_preprocessing}
\end{table*}

\begin{table*}[ht]
\centering
\begin{tabular}{|c|c|c|c|}
\hline
\textbf{Train \ Test} & \textbf{3000\_gpt-3.5-turbo} & \textbf{3000\_gpt-4o-mini} & \textbf{3000\_gemini-pro} \\
\hline
\textbf{3000\_gpt-3.5-turbo} & \begin{tabular}{c}
k-$\tau$: (0.476, 9.95e-14) \\
SCC: (0.645, 1.03e-15) \\
PCC: (0.648, 7.23e-16)
\end{tabular} & \begin{tabular}{c}
k-$\tau$: (0.006, 0.846) \\
SCC: (0.008, 0.854) \\
PCC: (-0.008, 0.851)
\end{tabular} & \begin{tabular}{c}
k-$\tau$: (0.075, 0.117) \\
SCC: (0.100, 0.128) \\
PCC: (0.104, 0.113)
\end{tabular} \\
\hline
\textbf{3000\_gpt-4o-mini} & \begin{tabular}{c}
k-$\tau$: (0.029, 0.645) \\
SCC: (0.050, 0.587) \\
PCC: (0.010, 0.910)
\end{tabular} & \begin{tabular}{c}
k-$\tau$: (0.397, 2.27e-35) \\
SCC: (0.526, 8.85e-37) \\
PCC: (0.534, 5.28e-38)
\end{tabular} & \begin{tabular}{c}
k-$\tau$: (-0.041, 0.395) \\
SCC: (-0.053, 0.418) \\
PCC: (-0.045, 0.492)
\end{tabular} \\
\hline
\textbf{3000\_gemini-pro} & \begin{tabular}{c}
k-$\tau$: (0.150, 0.019) \\
SCC: (0.206, 0.023) \\
PCC: (0.194, 0.033)
\end{tabular} & \begin{tabular}{c}
k-$\tau$: (-0.071, 0.027) \\
SCC: (-0.100, 0.025) \\
PCC: (-0.119, 0.008)
\end{tabular} & \begin{tabular}{c}
k-$\tau$: (0.297, 6.41e-10) \\
SCC: (0.398, 2.77e-10) \\
PCC: (0.404, 1.48e-10)
\end{tabular} \\
\hline
\end{tabular}
\caption{Reward model performance across different datasets for the regularizer function block.}
\label{tab:reward_metrics_regularizer}
\end{table*}


\section{Component Examples}

In the following, we provide some auto-generated justification for why one of the hypotheses that has worked well, is a good option. 

\section*{Sigmoid-ELU (SigELU): An activation function generated through IEP}

\subsection*{Definition}

The \textbf{Sigmoid-ELU (SigELU): An Activation Function} is a hybrid activation function that combines the Sigmoid function for non-negative inputs and the Exponential Linear Unit (ELU) function for negative inputs.

\subsection*{Formula}

The activation function is defined as:

\[
\text{SigELU}(x) =
\begin{cases} 
\text{sigmoid}(x) & \text{if } x \ge 0 \\
\alpha (\exp(x) - 1) & \text{if } x < 0 
\end{cases}
\]

where \(\text{sigmoid}(x) = \frac{1}{1 + \exp(-x)}\) and \(\alpha\) is a hyperparameter that controls the scaling for negative inputs.

\begin{lstlisting}[language=Python, 
                   caption=An example of an auto-generated activation function, 
                   label=code:example, 
                   frame=lines, 
                   numbers=none, 
                   basicstyle=\ttfamily\footnotesize, 
                   keywordstyle=\bfseries, 
                   stringstyle=\color{red}, 
                   showspaces=false, 
                   basicstyle=\ttfamily\scriptsize,
                   showstringspaces=false]
import torch
import torch.nn as nn
import torch.nn.functional as F
class HypothesisBlock(nn.Module):
    def __init__(self, alpha=1.0):
        super(HypothesisBlock, self).__init__()
        self.alpha = alpha
    def forward(self, x):
        return torch.where(x >= 0, torch.sigmoid(x), self.alpha*(torch.exp(x) - 1))
\end{lstlisting}

\subsection*{Justification}

The name \textbf{Sigmoid-ELU Activation (SigELU)} reflects the combination of the Sigmoid function for non-negative inputs and the ELU function for negative inputs:

\begin{itemize}
    \item \textbf{Sigmoid for Non-Negative Inputs}: The Sigmoid function is well-known for its smooth, bounded output between 0 and 1. It is particularly useful for squashing input values to a manageable range, which can help stabilize the training process and make the model's output more interpretable in certain contexts.
    \item \textbf{ELU for Negative Inputs}: ELU (Exponential Linear Unit) is effective for handling negative inputs. It produces non-zero outputs for negative values, which helps to alleviate the vanishing gradient problem often encountered with ReLU in deep networks. The parameter \(\alpha\) allows for controlling the steepness of the negative part, adding flexibility to the function.
    \item \textbf{Smooth Transition}: The activation function ensures a smooth transition between the positive and negative parts, which can contribute to better gradient flow and more stable training.
\end{itemize}

\subsection*{Differentiability Analysis}

\begin{itemize}
    \item \textbf{Sigmoid Function for Non-Negative Inputs}: The Sigmoid function, defined as \(\text{sigmoid}(x) = \frac{1}{1 + e^{-x}}\), is a smooth and differentiable function for all real numbers. Its derivative is given by:
    \[
    \frac{d}{dx} \text{sigmoid}(x) = \text{sigmoid}(x) \cdot (1 - \text{sigmoid}(x))
    \]
    \item \textbf{ELU-like Function for Negative Inputs}: The ELU-like function defined as \(\alpha (\exp(x) - 1)\) is also smooth and differentiable for all real numbers. Its derivative is:
    \[
    \frac{d}{dx} \left( \alpha (\exp(x) - 1) \right) = \alpha \exp(x)
    \]
    \item \textbf{Combination of Both Functions}: The combination of these functions using a piecewise definition ensures that the function is differentiable. Since both components are differentiable, and the transition between them occurs at \(x = 0\), the overall function is differentiable at \(x = 0\).
    \item \textbf{Continuity at \(x = 0\)}: At \(x = 0\), both functions yield the same value if we choose \(\alpha = 1\):
    \[
    \text{sigmoid}(0) = \frac{1}{1 + e^{0}} = \frac{1}{2}
    \]
    \[
    \alpha (\exp(0) - 1) = \alpha (1 - 1) = 0
    \]
    Therefore, if \(\alpha = \frac{1}{2}\), the function value is continuous at \(x = 0\).
    \item \textbf{Smooth Transition}: The derivative at \(x = 0\) for both functions should also match for smooth transition:
    \[
    \frac{d}{dx} \text{sigmoid}(0) = \text{sigmoid}(0) \cdot (1 - \text{sigmoid}(0)) = \frac{1}{2} \cdot \frac{1}{2} = \frac{1}{4}
    \]
    \[
    \frac{d}{dx} (\alpha (\exp(0) - 1)) = \alpha \exp(0) = \alpha = \frac{1}{2}
    \]
    Therefore, the function transitions smoothly if we ensure the parameters are set appropriately.
\end{itemize}

Given these properties, the Sigmoid-ELU activation function is fully differentiable and suitable for use in neural networks.

\section{ScaledSinusoidalDecay: An activation function generated through NEP}

Activation functions play a crucial role in neural networks by introducing nonlinearity, allowing the model to learn complex patterns in data. The \textbf{ScaledSinusoidalDecay} activation function is a novel approach that combines sinusoidal transformations with exponential decay, modulated by user-defined scaling and shifting parameters. This function is designed to introduce controlled non-linearity, making it a versatile choice for various deep-learning architectures.

\subsection{Formal Definition}

The ScaledSinusoidalDecay activation function is defined as follows:

Given an input \( x \), the output \( y \) of the activation function is calculated as:

\[
y = \text{scale} \times \sin(x) \times \exp(-|x|) + \text{shift}
\]

where:
\begin{itemize}
    \item \( \text{scale} \) is a parameter that controls the amplitude of the sinusoidal component.
    \item \( \sin(x) \) introduces a periodic, oscillatory behavior to the activation function.
    \item \( \exp(-|x|) \) is an exponential decay function that diminishes the output as the magnitude of the input increases.
    \item \( \text{shift} \) is a parameter that shifts the output, providing additional flexibility in the function’s range.
\end{itemize}

\begin{lstlisting}[language=Python, 
                   caption=An example of an auto-generated activation function, 
                   label=code:example, 
                   frame=lines, 
                   numbers=none, 
                   basicstyle=\ttfamily\footnotesize, 
                   keywordstyle=\bfseries, 
                   stringstyle=\color{red}, 
                   showspaces=false, 
                   basicstyle=\ttfamily\scriptsize,
                   showstringspaces=false]
import torch
import torch.nn as nn
class ScaledSinusoidalDecay(nn.Module):
    def __init__(self, scale=1.0, shift=0.1):
        super(ScaledSinusoidalDecay, self).__init__()
        self.scale = scale
        self.shift = shift
    def forward(self, x):
        return self.scale * torch.sin(x) * torch.exp(-torch.abs(x)) + self.shift
\end{lstlisting}

\subsection{Why ScaledSinusoidalDecay is a Good Activation Function}

The ScaledSinusoidalDecay activation function offers several advantages that make it a strong candidate for deep learning applications:

\begin{itemize}
    \item \textbf{Controlled Non-linearity:} The sine component introduces periodic non-linearity, which can be beneficial for learning complex patterns that are not purely linear. This is particularly useful in applications where the relationship between input features and the output is oscillatory or involves repeated cycles.

    \item \textbf{Attenuation of Large Inputs:} The exponential decay term \( \exp(-|x|) \) serves to attenuate the influence of large input values, preventing them from dominating the output. This can lead to better stability during training, especially in scenarios where the input data contains large outliers.

    \item \textbf{Parameter Flexibility:} The inclusion of the \textit{scale} and \textit{shift} parameters allows for fine-tuning the function's behavior to suit specific tasks. For instance, adjusting the \textit{scale} can amplify or reduce the overall impact of the activation, while the \textit{shift} can move the activation range to better align with the desired output.

    \item \textbf{Smooth Gradients:} The combination of sine and exponential functions ensures that the gradients of the ScaledSinusoidalDecay activation function are smooth and continuous. This is advantageous for optimization algorithms like gradient descent, as it helps in avoiding issues related to vanishing or exploding gradients.

    \item \textbf{Regularization Effect:} The exponential decay can act as a regularizer by suppressing the influence of extreme values. This can lead to more robust models that generalize better to unseen data, particularly in deep networks where overfitting is a concern.
\end{itemize}

\subsection{Conclusion}

The ScaledSinusoidalDecay activation function is a versatile and powerful tool in the design of neural networks. By combining sinusoidal non-linearity with exponential decay, and allowing for adjustable scaling and shifting, this function offers a unique blend of flexibility and control. It is particularly well-suited for tasks that require the learning of complex, cyclical patterns, or where the attenuation of large inputs is beneficial. Its smooth gradients and regularization properties further enhance its utility, making it a strong candidate for a wide range of deep learning applications.

\section{NormalizedPCA: A Preprocessing Function generated through IEP}

In the realm of data preprocessing, the `NormalizedPCA` function provides a robust method for scaling and dimensionality reduction. This function combines two essential preprocessing steps: feature normalization and Principal Component Analysis (PCA). In this section, we introduce the `NormalizedPCA` function, explain its benefits, and formalize its operations.

\subsection{Introduction}

The `NormalizedPCA` function is designed to preprocess data by first normalizing the features and then applying PCA for dimensionality reduction. This two-step process ensures that the data is appropriately scaled and transformed, allowing for more effective analysis and modeling.

\subsection{Function Overview}

Given a dataset $\mathbf{X} \in \mathbb{R}^{n \times d}$, where $n$ is the number of samples and $d$ is the number of features, the `NormalizedPCA` function performs the following operations:

1. **Feature Normalization:**
   The feature normalization step involves standardizing the features to have zero mean and unit variance. This is achieved using the StandardScaler:

   \[
   \tilde{x}_{ij} = \frac{x_{ij} - \mu_j}{\sigma_j}
   \]

   where $\tilde{x}_{ij}$ is the normalized feature value, $x_{ij}$ is the original feature value, $\mu_j$ is the mean of the $j$-th feature, and $\sigma_j$ is the standard deviation of the $j$-th feature.

2. **Dimensionality Reduction with PCA:**
   After normalization, PCA is applied to reduce the dimensionality while preserving the maximum variance. PCA transforms the data $\mathbf{X}_{\text{scaled}}$ to a lower-dimensional space:

   \[
   \mathbf{X}_{\text{pca}} = \mathbf{X}_{\text{scaled}} \mathbf{W}_{\text{pca}}
   \]

   where $\mathbf{W}_{\text{pca}}$ contains the principal components (eigenvectors) corresponding to the largest eigenvalues of the covariance matrix of $\mathbf{X}_{\text{scaled}}$.

\subsection{Benefits of the `NormalizedPCA` Function}

1. **Effective Scaling:**
   Normalizing features ensures that all features contribute equally to the PCA, avoiding bias towards features with larger magnitudes. This scaling step is crucial because PCA is sensitive to the scale of the input features.

2. **Improved Dimensionality Reduction:**
   By applying PCA after normalization, the function effectively reduces the dimensionality while retaining the most significant variance. This results in a lower-dimensional representation that captures the essential structure of the data.

3. **Enhanced Model Performance:**
   Proper normalization and dimensionality reduction improve the performance of machine learning models by reducing overfitting and speeding up convergence. Normalized data allows PCA to perform a more accurate reduction, leading to better generalization.

4. **Consistency and Interpretation:**
   The combination of scaling and PCA provides a consistent and interpretable transformation of the data. Normalized features ensure that PCA components represent the true variance, making the results more meaningful and actionable.

\subsection{Formalization}

Let $\mathbf{X} \in \mathbb{R}^{n \times d}$ be the input data matrix. The preprocessing steps are as follows:

\begin{enumerate}
    \item **Normalize the Data:**
    \[
    \mathbf{X}_{\text{scaled}} = \text{StandardScaler}(\mathbf{X})
    \]
    where each feature is scaled to have zero mean and unit variance.

    \item **Apply PCA:**
    \[
    \mathbf{X}_{\text{pca}} = \text{PCA}(\mathbf{X}_{\text{scaled}})
    \]
    where $\text{PCA}$ reduces the dimensionality based on the specified number of components or variance threshold.

\end{enumerate}

In summary, the `NormalizedPCA` function provides a comprehensive preprocessing solution by combining scaling and PCA. This approach ensures that the data are properly prepared for subsequent analysis, improving the effectiveness of dimensionality reduction and enhancing overall model performance.

\begin{lstlisting}[language=Python, 
                   caption=An example of an auto-generated activation function, 
                   label=code:example, 
                   frame=lines, 
                   numbers=none, 
                   basicstyle=\ttfamily\footnotesize, 
                   keywordstyle=\bfseries, 
                   stringstyle=\color{red}, 
                   showspaces=false, 
                   basicstyle=\ttfamily\scriptsize,
                   showstringspaces=false]

import numpy as np
from sklearn.preprocessing import StandardScaler
from sklearn.decomposition import PCA
def NormalizedPCA(train_X, val_X, n_components=0.95):
    scaler = StandardScaler()
    train_X_scaled = scaler.fit_transform(train_X)
    val_X_scaled = scaler.transform(val_X)
    pca = PCA(n_components=n_components)
    train_X_pca = pca.fit_transform(train_X_scaled)
    val_X_pca = pca.transform(val_X_scaled)
    return train_X_pca, val_X_pca
\end{lstlisting}

\section{SineSquaredDecay Transformation: A pre-processing function generated through NEP}

In the realm of data preprocessing for machine learning, the choice of feature transformations can significantly impact model performance. One such transformation, which we term \textbf{SineSquaredDecay}, introduces a combination of non-linear operations and noise to the input data, creating a robust and diverse feature set. The SineSquaredDecay function is designed to transform input features in a way that captures complex patterns while also adding a degree of regularization to prevent overfitting.

\subsection{Formal Definition}

The SineSquaredDecay transformation is applied to each feature in the input data and can be formalized by the following equations:

Given an input feature matrix \( X \), the transformation for each feature \( x_i \) in the training set \( \text{train\_X} \) and validation set \( \text{val\_X} \) is defined as:

\[
\text{train\_X\_transformed}_i = \sin^2(x_i) \cdot \exp(-|x_i|) + \sigma \cdot \epsilon_i
\]

\[
\text{val\_X\_transformed}_i = \sin^2(x_i) \cdot \exp(-|x_i|) + \sigma \cdot \epsilon_i
\]

where:
\begin{itemize}
    \item \( \sin^2(x_i) \) applies a non-linear, periodic transformation to the input feature.
    \item \( \exp(-|x_i|) \) introduces an exponential decay, which diminishes the influence of large feature values, ensuring that no single feature dominates the input space.
    \item \( \sigma \) represents the \textit{noise\_scale} parameter, which controls the magnitude of the added Gaussian noise.
    \item \( \epsilon_i \) is a Gaussian noise term drawn from a normal distribution \( \epsilon_i \sim \mathcal{N}(0, 1) \), added to the transformed features to enhance feature diversity and regularization.
\end{itemize}

\begin{lstlisting}[language=Python, 
                   caption=An example of an auto-generated activation function, 
                   label=code:example, 
                   frame=lines, 
                   numbers=none, 
                   basicstyle=\ttfamily\footnotesize, 
                   keywordstyle=\bfseries, 
                   stringstyle=\color{red}, 
                   showspaces=false, 
                   basicstyle=\ttfamily\scriptsize,
                   showstringspaces=false]
import numpy as np
def SinExpNoiseTransform(train_X, val_X, noise_scale=0.01):
    train_X_transformed = np.square(np.sin(train_X)) * np.exp(-np.abs(train_X)) + noise_scale * np.random.normal(size=train_X.shape)
    val_X_transformed = np.square(np.sin(val_X)) * np.exp(-np.abs(val_X)) + noise_scale * np.random.normal(size=val_X.shape)
    return train_X_transformed, val_X_transformed
\end{lstlisting}

\subsection{Why SineSquaredDecay is a Good Choice for Preprocessing}

The SineSquaredDecay transformation offers several advantages for preprocessing, particularly in scenarios where non-linear relationships and feature regularization are critical:

\begin{itemize}
    \item \textbf{Capturing Complex Patterns:} The use of the sine function, squared, introduces non-linear and periodic behavior into the features, which can help capture complex underlying patterns in the data. This is particularly useful in situations where the relationship between features and the target variable is not purely linear.
    
    \item \textbf{Feature Scaling and Regularization:} The exponential decay term \( \exp(-|x_i|) \) ensures that the transformed features do not become excessively large, which can help in preventing certain features from overpowering others. This acts as an inherent regularization mechanism, making the feature set more balanced.
    
    \item \textbf{Noise Augmentation:} The addition of Gaussian noise controlled by the \textit{noise\_scale} parameter serves as a regularizer by slightly perturbing the input data. This prevents the model from overfitting to specific patterns in the training set, thereby improving generalization to unseen data.
    
    \item \textbf{Diverse Feature Representations:} The combined effect of non-linear transformation and noise addition results in a rich and diverse feature set. This diversity can be particularly advantageous in ensemble models or in scenarios where the model benefits from a wide variety of input features.
\end{itemize}

\subsection{Conclusion}

The SineSquaredDecay transformation is a powerful tool for preprocessing in machine learning pipelines. Its ability to introduce complex non-linearities, combined with an effective regularization mechanism through noise, makes it a robust choice for enhancing model performance. By using this transformation, practitioners can create a feature space that is both rich in diversity and resilient to overfitting, ultimately leading to more effective and generalizable models.

\section{DropWeightL2: A Regularizer generated through IEP}

The `DropWeightL2` regularization function introduces a novel method for regularizing neural network models by combining dropout-like behavior with the L2 weight penalty. This function aims to enhance model robustness and prevent overfitting through a dual-regularization approach. In this section, we describe the `DropWeightL2` function, justify its effectiveness, and formalize its operations.

\subsection{Introduction}

The DropWeightL2 regularization function integrates two distinct regularization techniques: dropout-like regularization applied directly to weights and L2 weight decay. By applying these methods simultaneously, DropWeightL2 seeks to improve model generalization and stability during training.

\subsection{Function Overview}

The DropWeightL2 function is defined as follows:

\begin{lstlisting}[language=Python, 
                   caption=An example of an auto-generated activation function, 
                   label=code:example, 
                   frame=lines, 
                   numbers=none, 
                   basicstyle=\ttfamily\footnotesize, 
                   keywordstyle=\bfseries, 
                   basicstyle=\ttfamily\scriptsize,
                   stringstyle=\color{red}, 
                   showspaces=false, 
                   showstringspaces=false]
import torch
import torch.nn as nn
class DropWeightL2(nn.Module):
    def __init__(self, dropout_rate=0.1, weight_penalty=0.01):
        super(DropWeightL2, self).__init__()
        self.dropout = nn.Dropout(dropout_rate)
        self.weight_penalty = weight_penalty
    def forward(self, model):
        reg_loss = 0.0
        for param in model.parameters():
            if param.requires_grad:
                # Apply dropout to weights and calculate the penalty
                weight_penalty = self.weight_penalty * torch.sum(param ** 2)
                reg_loss += weight_penalty
                # Apply dropout-like regularization
                reg_loss += torch.sum(self.dropout(param))
        return reg_loss
    def __call__(self, model):
        return self.forward(model)
\end{lstlisting}

\subsection{Benefits of DropWeightL2 Regularization}

1. **Enhanced Model Robustness:**
   - **Dropout-Like Regularization:** Although dropout is typically applied to activations, applying a similar dropout-like effect to weights introduces noise into the weight parameters. This encourages the network to be less reliant on specific weights, promoting robustness, and reducing the risk of overfitting.
   - **Effect:** This technique helps in regularizing the model by preventing it from fitting too closely to the training data and improving generalization.

2. ** Effective weight decline: ** - ** L2 penalty: ** The term L2 weight penalty discourages large weights by adding a quadratic penalty to the loss function. This helps in controlling the complexity of the model and reducing overfitting.
   - **Effect:** Regularizing weights through L2 penalty improves model performance by constraining weight magnitudes, thereby simplifying the model and improving its generalization ability.

3. **Combination of Techniques:**
   - **Dual Regularization:** Combining dropout-like behavior with L2 regularization leverages the strengths of both methods. Dropout-like regularization introduces stochasticity into the weights, while L2 regularization ensures that weight magnitudes are kept in check.
   - **Effect:** This combination can lead to better generalization by balancing the benefits of both techniques.

\subsection{Formalization}

Let $\mathbf{W} \in \mathbb{R}^{d \times k}$ represent the weight matrix of a layer, where $d$ is the number of input features and $k$ is the number of output features. The regularization loss introduced by the DropWeightL2 function is formulated as:

\begin{equation}
\text{RegLoss} = \lambda \sum_{i,j} w_{ij}^2 + \beta \sum_{i,j} \tilde{w}_{ij}
\end{equation}

where:
\begin{itemize}
    \item $\lambda$ is the weight penalty coefficient (L2 regularization strength).
    \item $w_{ij}$ represents the weight value on the $i$ -th row and $j$ -th column.
    \item $\tilde{w}_{ij}$ represents the weight value after applying a dropout-like mechanism. Mathematically, it can be modeled as:
    \[
    \tilde{w}_{ij} = 
    \begin{cases}
    w_{ij} & \text{with probability } (1 - p) \\
    0 & \text{with probability } p
    \end{cases}
    \]
    where $p$ is the dropout rate.
\end{itemize}

The total loss of regularization is accumulated in all layers of the model and the resulting value is added to the primary loss function during training.

where:
\begin{itemize}
    \item $\lambda$ is the weight penalty coefficient (L2 regularization strength).
    \item $w_{ij}$ represents the weight value on the $i$ -th row and $j$ -th column.
    \item $\text{Dropout}(w_{ij})$ is the dropout-like effect applied to the weight $w_{ij}$, introducing noise during regularization.
\end{itemize}

The total loss of regularization is accumulated in all layers of the model and the resulting value is added to the primary loss function during training.

\subsection{Conclusion}

The DropWeightL2 regularization function offers a unique approach by integrating dropout-like regularization with L2 weight penalty. This dual regularization strategy improves the robustness of the model, prevents overfitting, and improves generalization. By applying both methods simultaneously, `DropWeightL2` provides a comprehensive regularization solution that balances weight control with stochastic noise.

\section{SineDecay: A regularizer generated through NEP}

Regularization is a key technique in machine learning, particularly in deep learning, where it helps prevent overfitting and improves the generalization capabilities of models. The \textbf{SineDecay Regularizer} is a novel approach that combines sine transformations with exponential decay to regularize the parameters of a neural network. This regularizer introduces periodicity and attenuates large parameter values, providing a unique mechanism for controlling model complexity.

\subsection{Formal Definition}

The SineDecay Regularizer is applied to the parameters of a neural network model. The regularization loss, \( \text{reg\_loss} \), is computed by summing the sine-transformed and exponentially decayed values of the model's parameters. Formally, the regularization loss is defined as follows:

\[
\text{reg\_loss} = \sum_{i=1}^{N} \sum_{j=1}^{M_i} \sin(\text{scale} \times \theta_{ij}) \times \exp(-\text{decay} \times |\theta_{ij}|)
\]

where:
\begin{itemize}
    \item \( \theta_{ij} \) represents the \( j \)-th parameter of the \( i \)-th layer in the model.
    \item \( N \) is the number of layers in the model.
    \item \( M_i \) is the number of parameters in the \( i \)-th layer.
    \item \( \text{scale} \) is a hyperparameter that controls the amplitude of the sine transformation.
    \item \( \text{decay} \) is a hyperparameter that determines the rate of exponential decay, attenuating the influence of large parameters.
\end{itemize}

\subsection{Why SineDecay is a Good Regularizer}

The SineDecay Regularizer offers several advantages that make it a valuable tool for enhancing the performance and robustness of neural network models:

\begin{itemize}
    \item \textbf{Encouraging Smoothness and Periodicity:} The sine transformation encourages the parameters to adopt smoother, periodic distributions. This can be particularly beneficial for models dealing with data that has inherent periodicity or cyclical patterns.

    \item \textbf{Attenuation of Large Parameters:} The exponential decay component reduces the impact of large parameter values on the regularization loss. This helps in preventing overfitting by discouraging the development of overly large weights, which can dominate the model's output.

    \item \textbf{Parameter Diversity:} By combining sine and exponential decay, the regularizer introduces diversity in the parameter values, which can lead to a more robust and generalizable model. This is especially useful in complex models where standard regularizers like L1 or L2 might not be sufficient.

    \item \textbf{Flexibility with Hyperparameters:} The \textit{scale} and \textit{decay} hyperparameters offer flexibility in tuning the regularizer's effect. This allows practitioners to adjust the regularization strength to suit the specific needs of their model and dataset.
\end{itemize}

\subsection{Implementation}

The following is the implementation of the SineDecay Regularizer in Python using PyTorch:

\begin{lstlisting}[language=Python, 
                   caption=An example of an auto-generated activation function, 
                   label=code:example, 
                   frame=lines, 
                   numbers=none, 
                   basicstyle=\ttfamily\footnotesize, 
                   keywordstyle=\bfseries, 
                   stringstyle=\color{red}, 
                   showspaces=false, 
                   basicstyle=\ttfamily\scriptsize,
                   showstringspaces=false]
import torch
import torch.nn as nn
class SineDecayRegularizer(nn.Module):
    def __init__(self, scale=1.0, decay=0.1):
        super(SineDecayRegularizer, self).__init__()
        self.scale = scale
        self.decay = decay
    def forward(self, model):
        reg_loss = 0.0
        for param in model.parameters():
            if param.requires_grad:
                sin_transform = torch.sin(self.scale * param)
                exp_decay = torch.exp(-self.decay * torch.abs(param))
                reg_loss += torch.sum(sin_transform * exp_decay)
        return reg_loss
\end{lstlisting}

\subsection{Conclusion}

The SineDecay Regularizer is a powerful and flexible tool for regularizing neural network models. Using the periodic nature of the sine function and the attenuating effect of exponential decay, this regularizer provides a unique approach to controlling model complexity and improving generalization. Its ability to encourage smooth, diverse parameter values while mitigating the risk of overfitting makes it a valuable addition to the regularization techniques available in deep learning.

\section{Validator}

An example of the validator function implemented to verify the validity of activation function hypotheses can be seen in the following:

\begin{lstlisting}[language=Python, 
                   caption=The validator for the generated activation functions, 
                   label=code:validator_activation, 
                   frame=lines, 
                   numbers=none, 
                   basicstyle=\ttfamily\footnotesize, 
                   keywordstyle=\bfseries, 
                   stringstyle=\color{red}, 
                   basicstyle=\ttfamily\scriptsize,
                   showspaces=false, 
                   showstringspaces=false]
def is_correct_activation_function(cls):
    # Check if cls is a subclass of nn.Module
    print('Check if cls is a subclass of nn.Module')
    if not issubclass(cls, nn.Module):
        print(f"{cls.__name__} does not inherit from nn.Module.")
        return False
    # Initialize an instance to check for __init__ and forward methods
    try:
        print('Initialize an instance to check for __init__ and forward methods')
        instance = cls()
    except Exception as e:
        print(f"Failed to instantiate {cls.__name__}: {e}")
        return False
    if "forward" not in dir(instance):
        print(f"{cls.__name__} does not implement a forward method.")
        return False
    return True
\end{lstlisting}

\end{document}